\documentclass[]{article}

\usepackage{amsmath}
\usepackage{csquotes}
\usepackage{multirow}
\usepackage{graphicx}
\usepackage[font=small,skip=2pt]{caption}
\usepackage{caption}
\usepackage{subcaption}
\usepackage{enumitem}
\usepackage[flushleft]{threeparttable}

\begin{document}

\title{Beyond Accuracy: ROI-driven Data Analytics of Empirical Data}

\author{Gouri Deshpande (gouri.deshpande@ucalgary.ca) \\ Guenther Ruhe (ruhe@ucalgary.ca) \\ Department of Computer Science, University of Calgary}
\date{}


\maketitle

\begin{abstract}
 \textit{Background:} 
The unprecedented access to data has rendered a remarkable opportunity to analyze, understand, and optimize the investigation approaches in almost all the areas of (Empirical) Software Engineering. However, data analytics is time and effort consuming, thus, expensive, and not automatically valuable. 
 
\textit{Objective:} 
This vision paper demonstrates that it is crucial to consider Return-on-Investment (ROI) when performing Data Analytics. Decisions on "How much analytics is needed"? are hard to answer. ROI could guide for decision support on the What?, How?, and How Much? analytics for a given problem.

\textit{Method:} The proposed conceptual framework is validated through two empirical studies that focus on requirements dependencies extraction in the Mozilla Firefox project. The two case studies are (i) Evaluation of fine-tuned BERT against Naive Bayes and Random Forest machine learners for binary dependency classification and (ii) Active Learning against passive Learning (random sampling) for \textit{REQUIRES} dependency extraction. For both the cases, their analysis investment (cost) is estimated, and the achievable benefit from DA is predicted, to determine a break-even point of the investigation.

\textit{Results:} For the first study, fine-tuned BERT performed superior to the Random Forest, provided that more than 40\% of training data is available. For the second, Active Learning achieved higher F1 accuracy within fewer iterations and higher ROI compared to Baseline (Random sampling based RF classifier). In both the studies, estimate on, How much analysis likely would pay off for the invested efforts?, was indicated by the break-even point.

\textit{Conclusions:} Decisions for the depth and breadth of DA of empirical data should not be made solely based on the accuracy measures. 
Since ROI-driven Data Analytics provides a simple yet effective direction to discover when to stop further investigation while considering the cost and value of the various types of analysis, it helps to avoid over-analyzing empirical data.
\\
\textbf{keywords}: Data Analytics, Return-on-Investment, Requirements Engineering, Dependency extraction, BERT, Mozilla
\end{abstract}


\section{Introduction}
Return-on-Investment (ROI) is of great interest in engineering and business for arriving at decisions. This is true in Software Engineering (SE) as well. For example, Silverio et al. \cite{martinez2013rearm} evaluated cost-benefit analysis for the adoption of software reference architectures for optimizing architectural decision-making. Cleland et al. \cite{cleland2004heterogeneous} studied the ROI of heterogeneous solutions for the improvement of the ROI of requirements traceability. Recent data explosion in the form of big data and advances in Machine Learning (ML) have posed questions on the efficiency and effectiveness of these processes that have become more relevant.
In this paper, we present a retrospective evaluation of two empirical studies taken from the field of requirements dependency analysis for the benefit of ROI.  

Data Analytics in SE (also called "Software Analytics" by Bird et al. \cite{bird2015art}) is a term widely used, sometimes with a slightly different meaning. We subsume all efforts devoted to collecting, cleaning, preparing, classifying, analyzing data, and interpreting the results as \textit{Data Analytics (DA)}. In SE, the goal of DA is to provide better insights into some aspects of the software development life-cycle, which could facilitate some form of understanding, monitoring, or improvement of processes, products or projects. 

\begin{figure}[!b]
\centering
\includegraphics[scale=0.3]{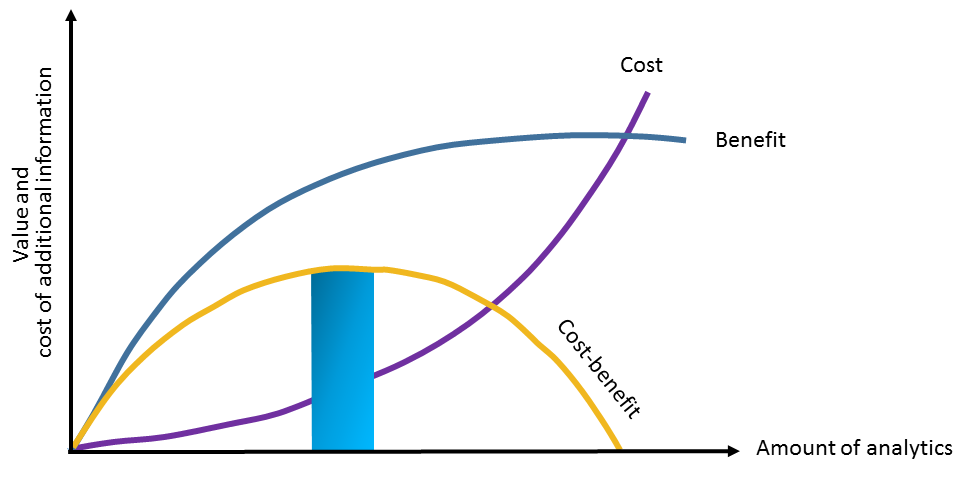}
\caption{Break-even point from cost-benefit analysis of technology investment.}
\label{fig:2}       
\end{figure}

SE is uncertain in various ways.  SE is highly human-centric, and processes are not strictly repeatable. The goals and constraints of software development are dynamically changing. Experimentation and DA are inherently arduous under such circumstances. The famous Aristotle \cite{barnes2004nicomachean} is widely attributed with a saying, "It is the mark of an educated mind to rest satisfied with the degree of precision which the nature of the subject admits and not to seek exactness where only an approximation is possible". Figure \ref{fig:2} shows a typical ROI (cost-benefit) curve of technology usage. Following some phase of increase, the curve reaches saturation, so, beyond that point, further investment does not pay off. We contemplate that a similar behaviour holds true for applying DA. \textbf{Our research hypothesis} is that ROI-driven DA helps to determine the break-even point of investment and thus optimizes resources spent in this process.
 
Paper structure: Section 2 discusses related work. The problem formulation is detailed in Section 3. 
Section 4 explains the empirical ROI investigation approach for the two problems.
A discussion of the applicability of the results is elaborated in Section 5. Finally, Section 6, provides an outlook on future research. 
\section{Related work}
\subsection{ROI Analysis in Software Engineering}
Evaluating the profitability of expenditure helps to measure success over a period of time thus takes the guesswork away from the concrete decision-making process.
For instance, Erdogmus et al.  \cite{erdogmus2004return} analyzed the  ROI  of quality investment to bring its importance in perspective and posed important questions,  \enquote{We generally want to increase a  software products quality because fixing existing software takes valuable time away from developing new software. But how much investment in software quality is desirable?  When should we invest, and where?}. 

Begel \& Zimmermann \cite{begel2014analyze} composed a set of 145 questions - based on a survey with more than 200 developers and testers - that are considered relevant for DA at Microsoft. One of the questions:\enquote{How important is it to have a software DA team answer this question?}, expected answer on a five-point scale (\textit{Essential} to \textit{I don't understand}). Although it provides a sneak peek of the development and testing environments of Microsoft, it does not prove any emphasis on any form of ROI. Essentially, we speculate that the ROI aspect was softened into asking for the perceived subjective importance through this question.

Boehm et al. \cite{boehm2008roi} presented quantitative results on the ROI of Systems Engineering based on the analysis of the 161 software projects in the COCOMO II database. Van Solingen \cite{van2004measuring} analyzed the ROI of software process improvement and took a macro perspective to evaluate corporate programs targeting the improvement of organizational maturity. Ferrari et al. \cite{ferrari2005roi} studied the ROI for text mining and showed that it has not only a tangible impact in terms of ROI but also an intangible benefits - which occur from the investment in the knowledge management solution that is not directly translated into returns, but that must be considered in the process of judgment to integrate the financial perspective of analysis with the non-financial ones. A lot of benefits occurring from the investment in this knowledge management solution are not directly translated into returns, but they must be considered in the process of judgment to integrate the financial perspective of analysis with the non-financial ones. 

Ruhe and Nayebi \cite{ruhe2016counts} proposed the \textit{Analytics Design Sheet} as a means to sketch the skeleton of the main components of the DA process. The four-quadrant template provides direction to brainstorm candidate DA methods and techniques in response to the problem statement and the data available. In its nature, the sheet is qualitative. ROI analysis goes further and adds a quantitative perspective for outlining DA.

\subsection{Empirical Analysis for Requirements Dependency Extraction}
The extraction of dependencies among requirements is an active field of SE research. 
The practical importance of the topic was confirmed by our survey \cite{deshpande_ruhe}. More than 80\% of the participants agreed or strongly agreed that (i) dependency type extraction is difficult in practice, (ii) dependency information has implications on maintenance, and (iii) ignoring dependencies has a significant ill impact on project success. 
 
In the recent past, many empirical studies have explored diverse computational methods that used natural language processing (NLP) \cite{och2002feasibility} \cite{atas2018automated}, semi-supervised technique \cite{deshpande_arora_ruhe_2019}, hybrid techniques \cite{deshpande_ruhe_2020} and deep learning \cite{guo2017semantically}. However, none of the approaches considered ROI to decide among techniques and the depth and breadth of their execution level. 

\section{Conceptual Framework for ROI-driven Data Analytics}
Different models exist that provide guidance to perform DA. Wieringa \cite{wieringa2014design} provides a checklist for what he calls the design cycle and the empirical cycle. In this study, we use the term \textit{Scoping} for defining the problem and the analysis objectives. Scoping also means defining the boundaries that help to exclude non-essential parts of the investigation. Analysis of the projected \textit{Return-on-Investment (ROI)} serves as an input for scoping.
\subsection{Research Question}
DA follows a resource and computation-intensive process constituting data gathering and processing components that are the non-trivial proportion of the total research cost. Thus, it is essential to account for these to compute the overall cost-benefit and optimize it further.

\textbf{Our aim is to look at DA} for empirical studies retrospectively (already conducted studies in the past). In particular, we are interested in Requirements Dependency Analysis (RDA) based studies. Through this research, we define and validate the principal concepts needed for ROI-driven DA. Our research question is:
\\ 

\textbf{RQ: What are the benefits of ROI-driven Data Analytics in the studies focusing on Requirements Dependency Analysis? }
 
\noindent \textbf{Justification:} As for any investment, it is most important to know how much is enough. There is no incentive to invest in analytics just for the sake of performing some analysis. Although one cannot claim exactness from this, it is worthwhile to get some form of guidance on where (which techniques) and how far (how much of it) one should go. To make the analysis concrete, we have selected RDA as the area of our specific investigations.
\begin{table}[!t]
\caption{Parameters used for ROI computation}
\renewcommand{\arraystretch}{1.1}
\begin{tabular}{p{1cm}| c p{3.25cm} c}

\hline
                                     & \textbf{Symbol}   & \textbf{Meaning}                               & \textbf{Unit} \\ \hline
\multirow{4}{*}{\textbf{Cost}} & $C_{dg}$          & Data gathering time                           & Minutes       \\  
                                     & $C_{pp}$          & Pre-processing time                             & Minutes       \\ 
                                     & $C_e$             & Evaluation time                              & Minutes       \\  
                                     & $C_l$             & Labeling time                                 & Minutes       \\ 
                                      & $C_{resource}$    & Human resource cost                            & \$ per hour         \\ 
                                     \hline \hline
\multirow{4}{*}{\textbf{Benefit}}    & $B_{reward}$      & Value per TP\ & \$          \\ 
                                     & $B_{penalty}$     & Penalty per FN \ & \$          \\   
                                     & $BF1_{iteration}$ & F1 difference                   & Number        \\
                                     & $PValue$          & Projected value per 1\% F1 improvement\ & \$           \\ \hline \hline
\multirow{5}{*}{\textbf{Others}}    
                                     & $H$               & \#Human resources                              & Number        \\   
                                     & $N_{train}$       & Size of the training set                          & Number        \\   
                                     & $N_{test}$        & Size of the test set                              & Number        \\  
                                     & $N$               & $N_{train}$ + $N_{test}$                       & Number        \\ \hline

\end{tabular}
\label{Params}
\end{table}

\subsection{Cost Factors}
\textbf{Data processing} is an umbrella term used to combine data collection ($C_{dg}$), pre-processing ($C_{pp}$) and labeling ($C_{l}$) under one hood, each one of which is a cost component. However, not all costs are fixed and some vary based on the solution approach used to tackle any decision problem. For example, supervised Machine Learning (ML) requires a large amount of annotated data, to begin with, whereas Active Learning acquires these annotations over a period of time in iterations until a stopping condition for classification operation is reached \cite{settles2009active}.  Additionally, there is a cost associated with modeling and evaluation ($C_{e}$). 

\subsection{Value Factors}
The value returns or \enquote{benefits} are defined based on the needs of the decision problem. In the context of dependency extraction, the benefit could be modeled in terms of the ability of the ML model to identify a larger number of dependencies correctly (higher \# of True Positives TP: $B_{reward}$) while limiting misclassification (reduced \# of False Negatives FN: $B_{penalty}$). Conversely, the benefit could also be determined based on the net value ($PValue$) of change of accuracy ($BF1_{iteration}$) in every iteration, especially when using Active Learning. 
Table \ref{Params} lists the relevant cost components and their corresponding units. These will be utilized to compute the $ROI$ later for the two different problems in Section 4.4.

\subsection{ROI}
To determine the ROI, we follow the simplest form of its calculation relating to the difference between $Benefit$ and $Cost$ to the amount of $Cost$. Both $Benefit$ and $Cost$ are measured as human effort in person hours.
\begin{equation} \label{eq:1}
\centering
ROI = (Benefit - Cost)/Cost 
\end{equation} 
Costa et al. \cite{costa2005intraoral} distinguished the “hard ROI” from the “soft ROI”. The former refers to the direct additional revenue generated and cost savings. The latter improved productivity, customer satisfaction, technological leadership, and efficiencies. 

\section{ROI of Techniques for Requirements Dependency Analysis}
We have selected the area of requirements dependency analysis (RDA) to illustrate and initially validate our former conceptual framework. In what follows, we introduce the key terms needed to formulate two Empirical Analysis Studies called EAS 1 resp. EAS 2.  

\subsection{Problem statement}
Following are the definitions of dependency types that are used to state the two studies. For a set of requirements $R$ and a pair of requirements $(r,s)$ $\epsilon$ $R\times R$
\begin{itemize}[leftmargin=*]
  \item[1)]  An \textbf{\textit{INDEPENDENT}} relationship is defined as the absence of any form of relationship between a pair of requirements.
 \item [2)]  A \textbf{\textit{DEPENDENT}} relationship is defined as the complement set of INDEPENDENT.  i.e., there exists at least one type of the dependency types such as
\textit{REQUIRES, SIMILAR, OR, AND, XOR, value synergy,
effort synergy} etc. between $r$ and $s$.
 \item [3)] \textbf{\textit{REQUIRES}} is a special form of \textit{DEPENDENT} relationship.  If $r$ requires $s$, or $s$ requires $r$, then, $r$ and $s$ are in a \textit{REQUIRES} relationship  
 \item [4)] \textbf{\textit{OTHER}} type of dependency is when $(r,s)$ is \textit{DEPENDENT} and the dependency type is not REQUIRES (could be any of the other dependency types mentioned in (2)) 

\end{itemize}

\begin{enumerate}[wide, labelwidth=!, labelindent=0pt]
 \item  [\textbf{Problem 1-}] \textbf{ Binary requirements dependency extraction:}
For a given set $R$ of requirements and their textual description, the binary requirements dependency extraction problem aims to classify each pair (r,s) $\epsilon$ $R\times R$ as \textit{DEPENDENT} or \textit{INDEPENDENT}. \\
 \item [\textbf{Problem 2-}] \textbf{Specific requirements dependency extraction of the type \textit{REQUIRES}:}
For a given set $R$ of requirements and their textual description, the REQUIRES dependency extraction problem aims to classify for each pair (r,s) $\epsilon$ $R\times R$ if they are in a \textit{$REQUIRES$} relationship.
  
\end{enumerate}

\subsection{Empirical Analysis Studies (EAS)}
In this section, we formulate two Empirical Analysis Studies, EAS 1 and EAS 2, to investigate the two problems explained above. We aim to analyze and compare Bidirectional Encoder Representations from Transformers (BERT), and Active Learning (AL), both proven to be of interest in general and pre-evaluated for their applicability to the stated problems, with traditional ML. For the two studies, we examine the (F1) accuracy and the ROI of the whole process of DA. 

\textbf{EAS 1:} We compare two supervised classification algorithms: Naive Bayes (NB) and Random Forest (RF) - ML algorithms successfully and prominently used for text classification\cite{manning2010introduction} in the past, with a fine-tuned BERT model \cite{devlin2018bert}. The analysis was performed for an incrementally growing training set size to capture its impact on F1 accuracy and ROI.

\textit{BERT} (Bidirectional Encoder Representations from Transformers) \cite{devlin2018bert} is a recent technique published by researchers from Google. BERT is applying bidirectional training of Transformer, a popular attention model, to language modeling, which claims to be state-of-the-art for NLP tasks. In this study scenario, we explore the question, \enquote{How does fine-tune BERT compare with traditional algorithms on an economical scale?} by comparing models' effectiveness with incurred ROI. 

\textbf{EAS 2:} Random sampling (Passive Learning) randomly selects a training set - referred to as \textit{Baseline} in the rest of the paper. Active Learning selects the most informative instances using various sampling techniques such as MinMargin and LeastConfidence \cite{settles2009active}. We compare \textit{Baseline} with AL using RF as a classifier for this scenario. The analysis was done by adding a few training samples in every iteration concurrently to classify the unlabeled instances.

\textit{Active Learning} (AL) is a ML method that guides a selection of the instances to be labeled by an oracle (e.g., human domain expert or a program) \cite{settles2009active}. While this mechanism has been proven to positively address the question, \enquote{Can machines learn with fewer labeled training instances if they are allowed to
ask questions?}, through this exploration, we try to answer the question,\enquote{Can machines learn more economically if they are allowed to ask questions?} \cite{settles2011theories}.

\subsection{Data}
The online bug tracking system Bugzilla \cite{bugzilla} is widely used in open-source software development. New requirements are logged into these systems in the form issue reports \cite{shi2017understanding} \cite{bhowmik2015resolution} which help software developers to track them for effective implementation \cite{shin2015guidelines}, testing,  and release planning. In Bugzilla, feature requests are a specific type of issue that is typically tagged as “enhancement” \cite{mozillawiki}. We retrieved these feature requests or requirements from \textit{Firefox} and exported all related fields such as Title, Type, Priority, Product, Depends\_on, and See\_also. 

\textbf{Data collection:} Collecting data from Bugzilla was a substantial effort that was carried out in multiple rounds. We collected 3,704 enhancements from \textit{Firefox} using REST API through a python script such that each one of the enhancements considered for retrieval is dependent on at least another one in the dataset. The data spanned from 08/05/2001 to 09/08/2019.

\textbf{Data preparation:} The complete data was analyzed to eliminate special characters and numbers. Then dependent requirement pairs were created based on the depends\_on (interpreted as \textit{REQUIRES} dependency) field information for each one of the enhancements. Requirements with no dependency between them were paired to generate \textit{INDEPENDENT} class dataset. Further, sentence pairs that had fewer than three words in them were filtered out resulting in 3,373 \textit{REQUIRES}, 219 \textit{OTHER} and 21,358 \textit{INDEPENDENT} pairs. 

\textbf{Pre-processing and feature extraction:} The data was first processed to eliminate stop words and then lemmatized following the traditional NLP pipeline \cite{arellano2015frameworks}. For supervised and AL ML, we used the Bag Of Words (BOW) \cite{ramos2003using} feature extraction method, which groups textual elements as tokens. For applying BERT, we retained sentence pairs in their original form (without stop word removal and lemmatization).

\textbf{Classifiers:} For both NB and RF, the data was split into train and test (80:20) and balanced between classes. Also, hyper-parameter tuning was performed and the results for 10-fold cross-validation were computed, followed by testing (on unseen data).
\par To fine-tune the BERT model, we used \textit{NextSentencePrediction}\footnote{https://huggingface.co/transformers/model\_doc/bert.html\#bertfornextsentenceprediction}, a sentence pair classification pre-trained BERT model, and further fine-tuned it for the RDA specific dataset on Tesla K80 GPU on Google Colab\footnote{https://colab.research.google.com/}.

\subsection{ROI Modeling}
\subsubsection{\textbf{EAS1}}
The classification algorithms such as RF and NB, have been explored in NLP based SE problems. These algorithms are driven by the feature extraction aspect to a great extent. Thus, could influence their effectiveness on classification outcomes. 
However, feature extraction is problem specific and incurs substantial cost and access to domain expertise.
\par On the other hand, BERT eliminates the need for feature extraction since it is a language model based on deep learning. BERT, pre-trained on a large text corpus, can be fine-tuned on specific tasks by providing only a small amount
of domain-specific data. 

In this empirical analysis, we conducted classification by utilizing a fraction of the whole dataset for training and testing for a small fixed data set. This was repeated by slowly increasing the fraction of the training set and results were captured. 

\par During every classification, $Cost$ and $Benefit$ were computed using various parameters explained in Table \ref{Params}. $Cost$ is the sum of the data processing costs ($(C_{dg} + C_{pp} +  C_e + C_l)/60$) (in hours) for a fraction (N\%) of training set. This is further translated into dollar cost based on hourly charges ($C_{resource}$) of $H$ human resources. 
\begin{equation} \label{eq:2}
    Cost = N\% *  \frac {(C_{dg} + C_{pp} +  C_e + C_l)}{60}*H* C_{resource}
\end{equation}
$Return$ computations for RDA, assumes reward ($B_{reward}$) for identifying the dependent requirements (TP) while penalizing ($B_{penalty}$) instances that were falsely identified as independent (FN).
\begin{equation} \label{eq:3}
     Benefit = TP*B_{reward} - FN*B_{penalty}
\end{equation}

\begin{table}[!htpb]
\centering
\caption{Parameter settings for the two empirical analysis scenarios}
\renewcommand{\arraystretch}{1.18}
\begin{threeparttable}[t]
  \centering
\begin{tabular}{p{4cm}    p{3cm}}
\hline
\textbf{Parameters} & \textbf{Values}\\ \hline 
$C_{fixed} = C_{dg} + C_{pp} +  C_e $ & 1 min/sample                         \\  
$C_l$                                 & 0.5 min/sample                       \\  
$C_{resource}$                        & \$400/hr                             \\  
$H$                                   & 1                                    \\ 
$N$                                   & 4,586  
\\ \hline 
$B_{reward}$                          & \$500/TP                             \\  
$B_{penalty}$                         & \$500/FN                             \\  
$BF1_{iteration}$                     & =$F_{cur} - F_{prev}$                \\ 
$PValue$                              & \$10,000 per percent F1 improvement \\ \hline                   
\end{tabular}
\end{threeparttable}
\label{Assumptions}
\end{table}

\subsubsection{\textbf{EAS 2}}
In this empirical analysis, we compared AL with a traditional random sampling based classification- \textit{Baseline} - using the RF ML algorithm. 
\par Beginning with 60 training samples of each class (\textit{REQUIRES, INDEPENDENT and OTHER)}, we developed multi-class classifiers for both AL and Baseline for this empirical study scenario. When AL used MinMargin sampling technique\footnote{MinMargin sampling technique performed well compared to Least Confidence and Entropy thus, we utilized MinMargin for this study} to identify 20\footnote{The tests were performed with\#samples = 10, 15 and 20. In this study, we will discuss results related to \#samples=20} most uncertain instance (requirement pair) for oracle to label, baseline randomly selected 20 instances and added to the training set along with their label, thus, kept the two approaches comparable in all the 20 iterations. Since data is already labeled, for AL, we pretend they are unlabeled until queried and labeled by a simulated oracle in this scenario.

\par The  $Cost$ is determined by first computing the sum of total processing time in person hours (= $Cost$) taken for data processing ($C_{fixed} = C_{dg} + C_{pp} + C_e)$), labeling ($C_l$) of train set ($N_{train}$) and data processing cost ($C_{fixed}$) for testing. This is further translated into dollar cost (=$C_{total}$) based on hourly charges ($C_{resource}$) of $H$ human resources.
\begin{equation*} \label{eq:7A}
    Cost = \frac {N_{train} * (C_{fixed} + C_l) + N_{test}*C_{fixed}}{60}
\end{equation*}
\begin{equation} \label{eq:4}
    C_{total} = Cost* H * C_{resource}
\end{equation}
\par Likewise, $Benefit$ is defined as the monetary value associated with a 1\% improvement in F1 score ($BF1_{iteration}$) between subsequent iterations. 

\begin{equation} \label{eq:5}
    Benefit = BF1_{iteration} * PValue
\end{equation}

\section{Results}
\begin{figure}[!t]
    \centering
        \includegraphics[scale=.32]{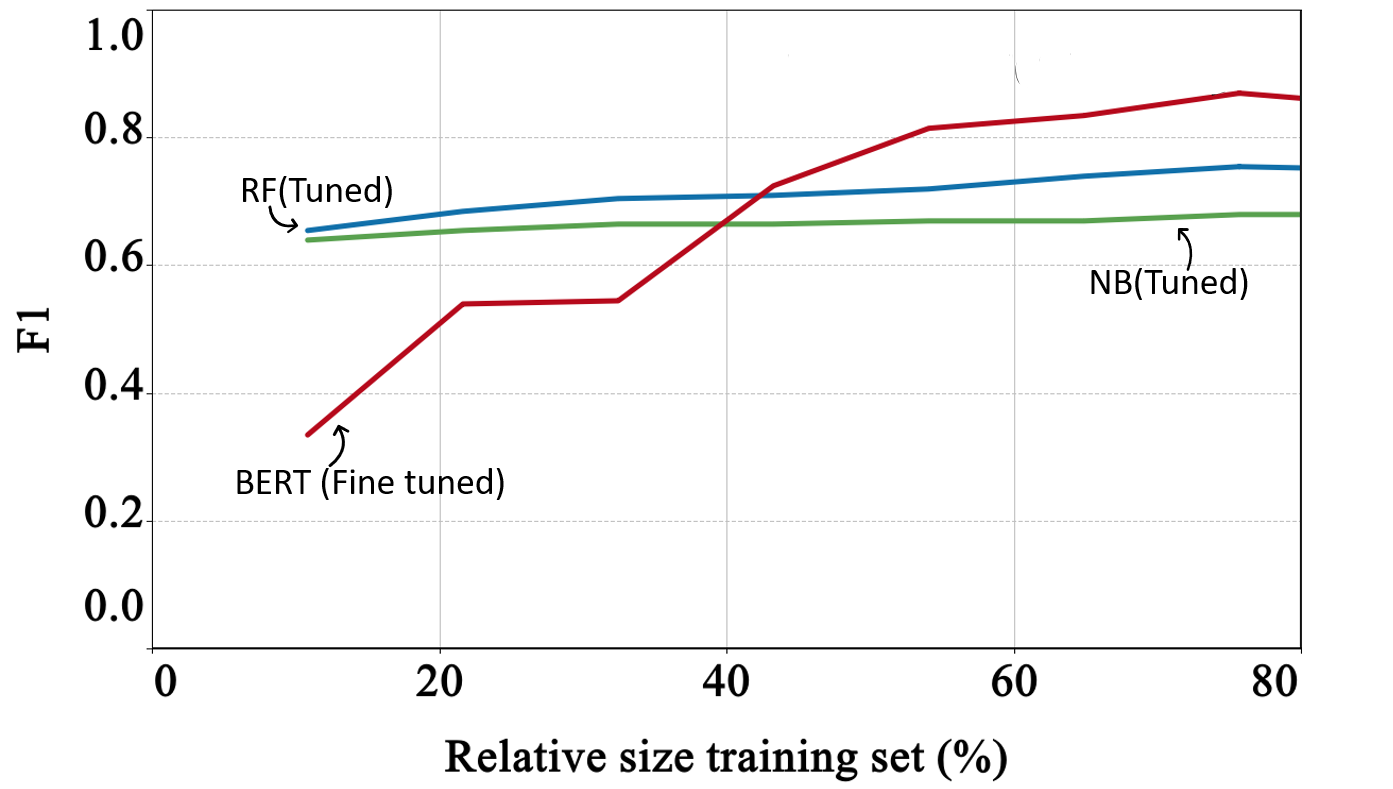}
        \caption{F1 score plot for NB, RF and BERT trained over increasing training set size, F1 improves, but plateaus beyond a certain point}
        \label{fig:BERTvsOthers}
\end{figure}

In the real-world, cost and benefit values are hard to get and are uncertain. All the results presented in this section are based on the parameter settings given in Table \ref{Assumptions}. The settings reflect practical experience but are not taken from a specific data collection procedure. We claim that the principal arguments made in our paper are independent of these settings. 

\subsection{EAS 1}
\begin{figure}[hbt!]
\centering
\begin{subfigure}[t]{.47\linewidth}
  \centering
    \includegraphics[width = 4.1cm, height = 3cm]{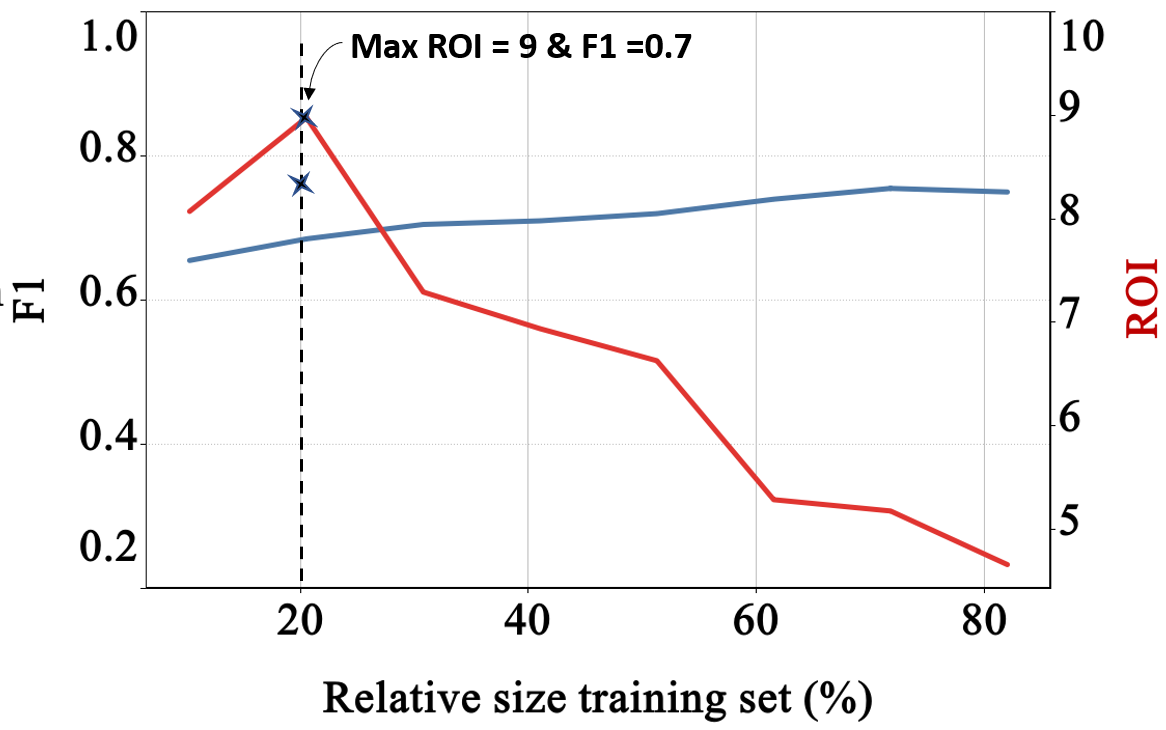}
        \caption{ F1 vs ROI for Random Forest}
        \label{fig:RF_ROI}
\end{subfigure}
\hfill
\begin{subfigure}[t]{.47\linewidth}
  \centering
\includegraphics[width = 4cm, height = 3cm]{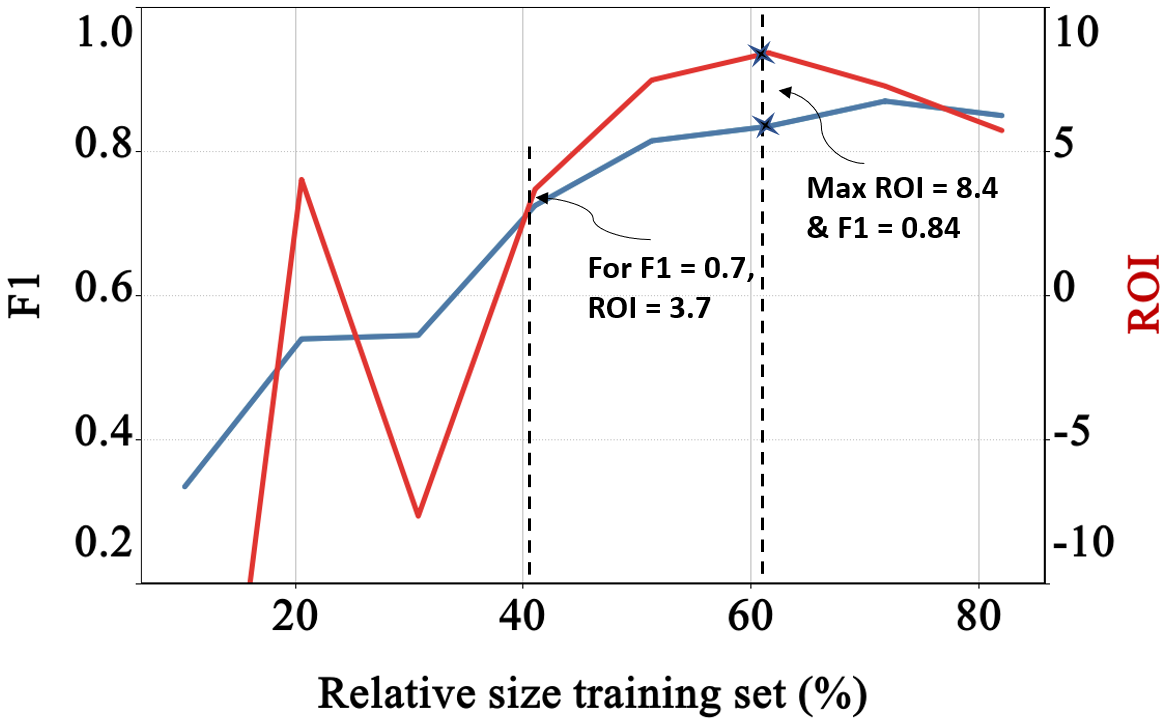}
        \caption{ F1 vs ROI for Fine tuned BERT }
        \label{fig:BERT_ROI}
\end{subfigure}
\caption{Empirical Analysis Scenario 1 (EAS 1)}
\label{fig:EAS1}
\end{figure}

Figure \ref{fig:BERTvsOthers} provides the \enquote{accuracy only view} and shows that F1 gradually increases with the increasing training size for the three ML algorithms: NB, RF, and BERT. However, all three ML algorithms reach a saturation towards larger training set sizes. While BERT performed exceptionally well when training set size exceeded 42\%, it could have been ideal to pre-determine \enquote{How much training is enough?}. Thus we selected the top two classifiers (Figure \ref{fig:BERTvsOthers}): BERT and RF and applied the monetary values (Table \ref{Assumptions}) for the various cost and benefit factors defined in Table \ref{Params} and computed the ROI.

\par Figure \ref{fig:RF_ROI} and \ref{fig:BERT_ROI} show the results for RF and BERT, respectively. The ROI behaviour is not monotonous and peaks for both cases. Although RF classification achieved the highest ROI with just 20\% of training set and accuracy of F1 = 0.7, highest F1 value of 0.75 was achieved along with the lowest ROI of 4.7. 

 \textbf{   For RF classification and applying ROI arguments, learning can be stopped with 20\% of the training set.}
    
Now looking at BERT classification, the best ROI-driven results: F1 = 0.84 and an ROI = 8.43, were achieved with the 60\% training set. Although F1 rose to 0.9 with 70\% training set size, ROI dropped to 7.27. For the recommendation of 20\% of training set size, ROI has a local optimum. BERT in general performs well on the F1, however, is it worth the ROI? needs to be explored. 

    \textbf{For training set sizes of at least 40\% of the size of the whole set, BERT performed better than RF in terms of both accuracy and ROI.}

\subsection{EAS 2}
\begin{figure}[!t]
\centering
\begin{subfigure}[t]{0.45\linewidth}
\centering
    \includegraphics[height= 3cm, width = 4cm]{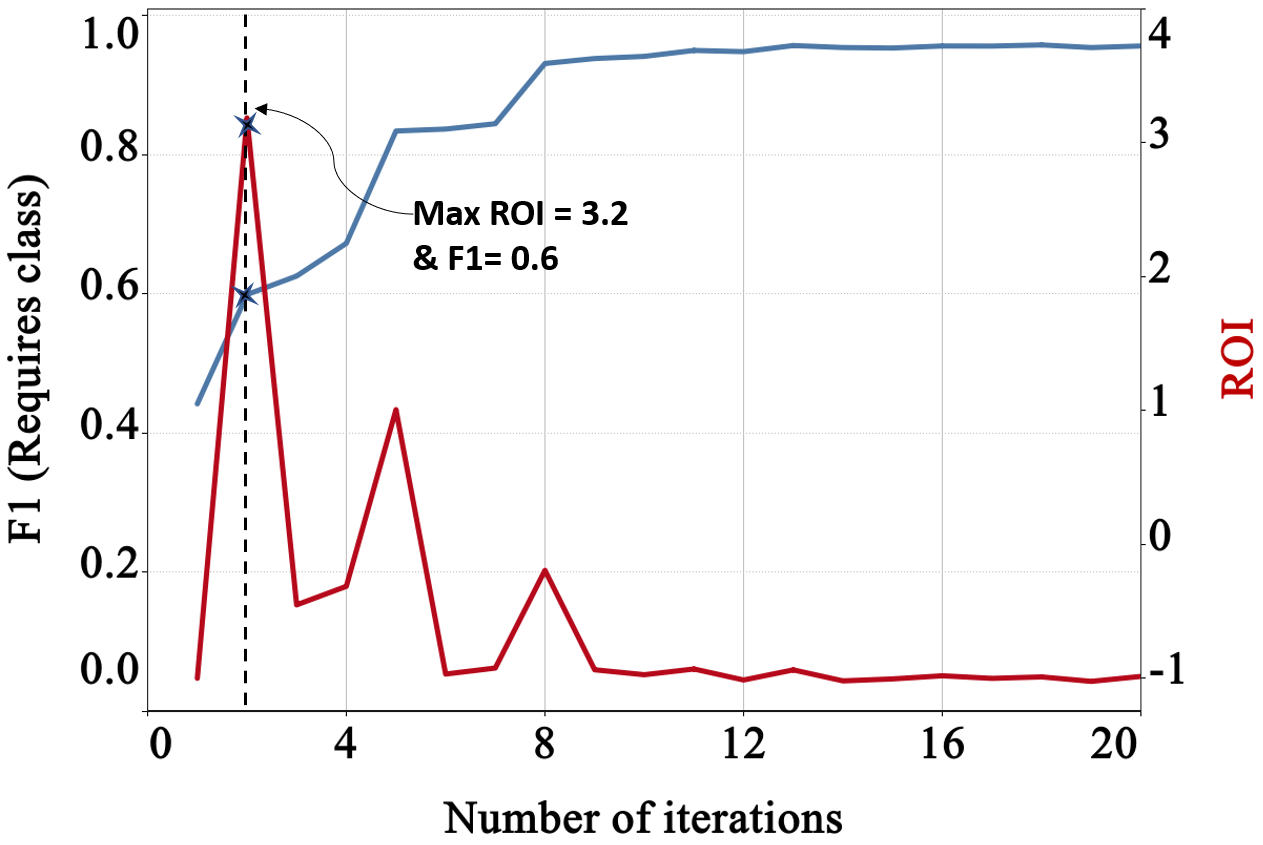}
        \caption{F1 vs ROI for Baseline}
        \label{fig:Baseline_ROI}
  \end{subfigure}\hfill
  \begin{subfigure}[t]{0.45\linewidth}
  \centering
\includegraphics[height= 3cm, width = 4cm]{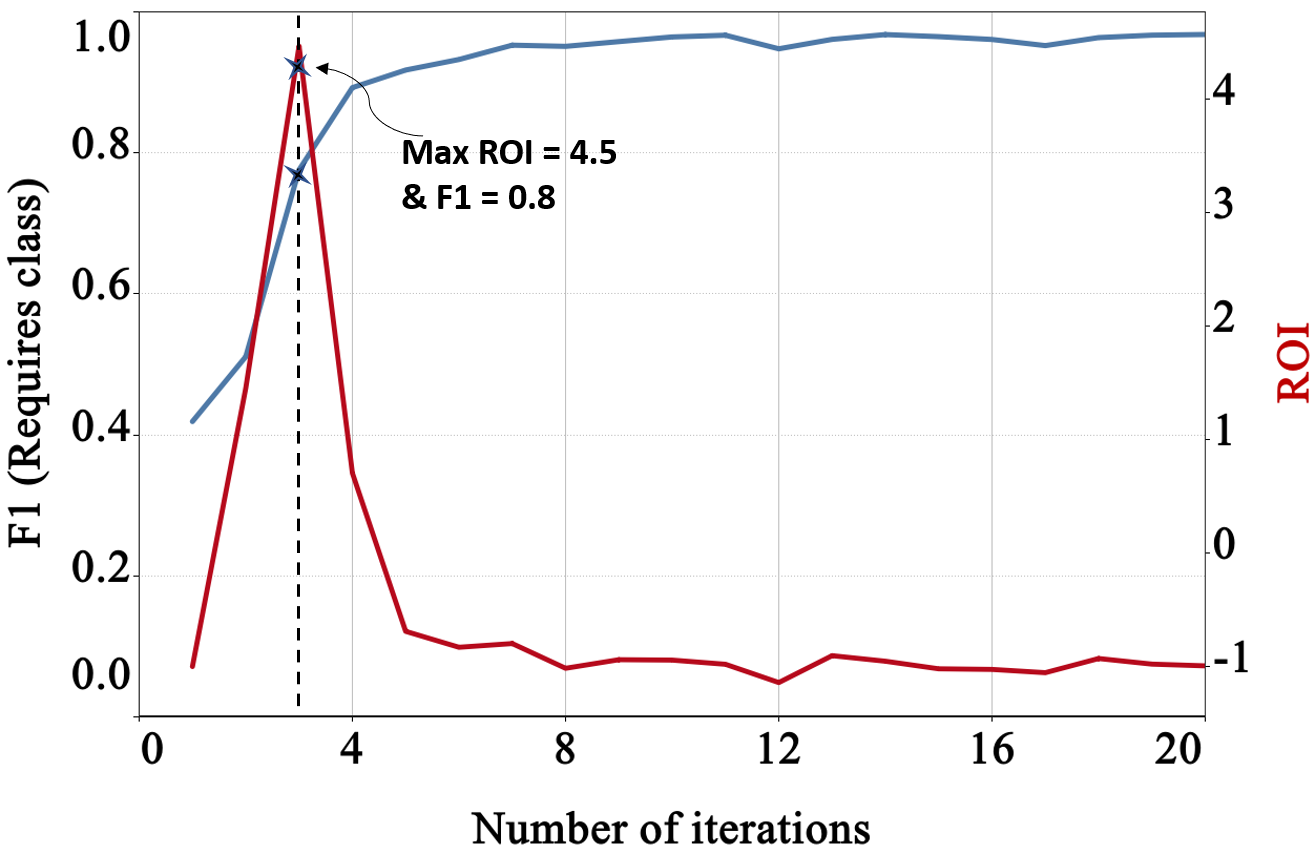}
        \caption{ F1 vs ROI for AL}
        \label{fig:AL_ROI}
  \end{subfigure}
\caption{Empirical Analysis Scenario 2 (EAS2)}
\label{fig:EAS2}
\end{figure}

We analyzed the ROI for \textit{Baseline} against AL for classifying the \textit{REQUIRES} class. The results are shown in Figure \ref{fig:Baseline_ROI} and Figure \ref{fig:AL_ROI}. Similar to EAS 1, we applied the values from Table \ref{Assumptions} and equations (\ref{eq:4}) and (\ref{eq:5}) to compute cost and benefit at every iteration for both the approaches. For the \textit{Baseline} approach, ROI peaked at 3.2 and F1 = 0.6, in the very 2nd iteration. Onwards, ROI drastically decreased which indicated lesser value for increasing training set by random sampling (\textit{Baseline}) method. 

Similar behavior was observed for the AL approach. shown in Figure \ref{fig:AL_ROI}. The peak here was after three iterations with values ROI = 4.5 and F1 = 0.8. 

\textbf{Both Baseline and AL showed the best ROI performance in the early iterations. Higher F1 accuracy needs additional human resources and reduces the ROI.}

\section{Discussion} 
For the problem of RDA, we explored the potential value of ROI-driven decisions. When chasing higher accuracy, there is a risk of over analyzing empirical data. In the sense that the value added due to increased accuracy is not justifiable by the additional effort needed in achieving it. 
\par \textbf{What does a high or low ROI mean for DA? : }
If available, a high ROI ratio indicates that there is a substantial benefit expected from following the recommendations derived from DA. Assuming that the ROI-driven suggestions are implemented, the small improvements achieved for solving the decision problems with high impact could justify the effort invested. Analysis related to effort and benefit, targeting high ROI, also implies simplicity first. Advanced methods are needed, but they are hard to justify practical application if a similar type of insight could be reached from a much simpler analysis, e.g., from descriptive statistics.
\par \textbf{What is the risk of ignoring  analysis?: }
The calculation of ROI is based on the value and effort estimates and thus only provides an approximation. In all types of exploratory data analysis, the emphasis is mainly on creating new research hypotheses or validating existing assumptions. In these cases, the notion of ROI is not the primary concern. Also, estimates for value and effort needed are highly dependent; hence, the ROI might only serve as a soft recommendation. On the other hand, whenever the ROI can be determined as a reasonable estimate, even after using intervals of best and worst-case performances, then ignoring ROI means to potentially waste effort for analysis that does not pay off the investment made. For EAS 1, if the training size set was limited to 30\%, RF could be considered as a better choice over BERT. However, with the possibility to increase the training set size, the BERT approach could be favored.

\section{Conclusions and Future Work}
We proposed to complement Data Analytics of empirical studies with ROI analysis to avoid over analyzing data in this vision paper. To validate the need, we performed an analysis of accepted papers of ESEM conferences between 2015 and 2019 and found that 51 out of 190 papers (27\%) were addressing some form of DA. Among them, 39\% included some consideration of cost, value, or benefit. However, none of them directly explored or discussed ROI or used cost-benefit analysis to decide the degree of DA needed. From a decision-making perspective, selecting one out of many techniques, and for a selected technique, deciding the termination of analysis amount to enlarge the scope from one to two criteria. 
\par Beyond accuracy, reflecting the benefit, it is essential to look into the investment as well. Exclusively looking into the different aspects of accuracy is cardinal, but it does not provide a full picture as the effort consumption and impact are ignored. Effort estimation is well studied, however; prediction of value \cite{biffl2006value} has not been explored as much. Even rough estimates may be helpful to decide how much further investment into DA is reasonable. To make this agenda successful, economical, business, and social 
 concepts need to be taken into account, apart from just the technical aspects.
 
\section*{Acknowledgement}
We thank Atharva Naik and Venessa Chan for useful comments. This work is supported by the Natural Sciences and Engineering Research Council of Canada, Discovery Grant RGPIN-2017-03948. 
\bibliographystyle{abbrv}
\bibliography{./main.bbl}

\end{document}